\documentclass[conference]{IEEEtran}
\IEEEoverridecommandlockouts
\usepackage{cite}
\usepackage{amsmath,amssymb,amsfonts}
\usepackage{graphicx}
\usepackage{textcomp}
\usepackage{xcolor}
\usepackage{footnote}
\usepackage{url}
\usepackage{xurl}
\usepackage{booktabs}
\usepackage{multirow}
\usepackage{multicol}
\usepackage{arydshln}
\newcommand{\norm}[1]{\left\lVert#1\right\rVert}
\usepackage{pifont}
\newcommand{\cmark}{\ding{51}}%
\newcommand{\xmark}{\ding{55}}%

\RequirePackage{algorithm}
\RequirePackage{algorithmic}

\RequirePackage{etoolbox}
\definecolor{cb-black}      {RGB}{  0,   0,   0}
\definecolor{cb-blue-green} {RGB}{  0,  073,  073}
\definecolor{cb-green-sea}  {RGB}{  0, 146, 146}
\definecolor{cb-rose}       {RGB}{255, 109, 182}
\definecolor{cb-salmon-pink}{RGB}{255, 182, 119}
\definecolor{cb-purple}     {RGB}{ 73,   0, 146}
\definecolor{cb-blue}       {RGB}{ 0, 109, 219}
\definecolor{cb-lilac}      {RGB}{182, 109, 255}
\definecolor{cb-blue-sky}   {RGB}{109, 182, 255}
\definecolor{cb-blue-light} {RGB}{182, 219, 255}
\definecolor{cb-burgundy}   {RGB}{146,   0,   0}
\definecolor{cb-brown}      {RGB}{146,  73,   0}
\definecolor{cb-clay}       {RGB}{219, 209,   0}
\definecolor{cb-green-lime} {RGB}{ 36, 255,  36}
\definecolor{cb-yellow}     {RGB}{255, 255, 109}

\def\BibTeX{{\rm B\kern-.05em{\sc i\kern-.025em b}\kern-.08em
    T\kern-.1667em\lower.7ex\hbox{E}\kern-.125emX}}
\begin{document}

\title{Time Series Anomaly Detection using
Diffusion-based Models
}

\author{\IEEEauthorblockN{Ioana Pintilie}
\IEEEauthorblockA{\textit{Bitdefender} \\
\textit{University of Bucharest}\\
Bucharest, Romania \\
ipintilie@bitdefender.com}
\and
\IEEEauthorblockN{Andrei Manolache}
\IEEEauthorblockA{\textit{Bitdefender} \\
\textit{University of Stuttgart}\\
Stuttgart, Germany \\
amanolache@bitdefender.com}
\and
\IEEEauthorblockN{Florin Brad}
\IEEEauthorblockA{\textit{Bitdefender}
\\
\textit{}\\
Bucharest, Romania \\
fbrad@bitdefender.com}}

\maketitle

\begin{abstract}
Diffusion models have been recently used for anomaly detection (AD) in images. In this paper we investigate whether they can also be leveraged for AD on multivariate time series (MTS). We test two diffusion-based models and compare them to several strong neural baselines. We also extend the PA\%K protocol, by computing a $ROC_K$-AUC metric, which is agnostic to both the detection threshold and the ratio $K$ of correctly detected points. Our models outperform the baselines on synthetic datasets and are competitive on real-world datasets, illustrating the potential of diffusion-based methods for AD in multivariate time series.
\end{abstract}


\section{Introduction}
Anomaly Detection (AD) in Multivariate Time Series (MTS) is an important research topic concerned with detecting irregular patterns in multidimensional temporal data. Developing such detectors is a critical element in predictive maintenance for Industry 4.0 sectors~\cite{Zonta2020} such as manufacturing or IT infrastructure, enabling the continuous monitoring of  sensors from different equipments.

Due to the scarcity of labeled training data, most of the methods are unsupervised or semi-supervised. Classical methods for AD such as OC-SVM~\cite{Scholkopf2001} or IsolationForest~\cite{Liu2008} do not explicitly handle temporal information.
On the other hand, deep learning methods for MTS model the temporal dimension. Many of them fall under the \emph{reconstruction-based} approaches, which learn features of normality with an autoencoding objective, then use the deviation of an example's reconstruction as an anomaly score. Some of these approaches involve generative models (such as GANs~\cite{Li2023,Li2019,Zhou2019}) to learn to generate normal time series adversarially. Recent works use diffusion models~\cite{Zhang2023,Wyatt2022}, which have superior mode coverage to GANs and improve AD in images. 

\begin{figure}[t!]
    \centering
    \includegraphics[width=\linewidth]{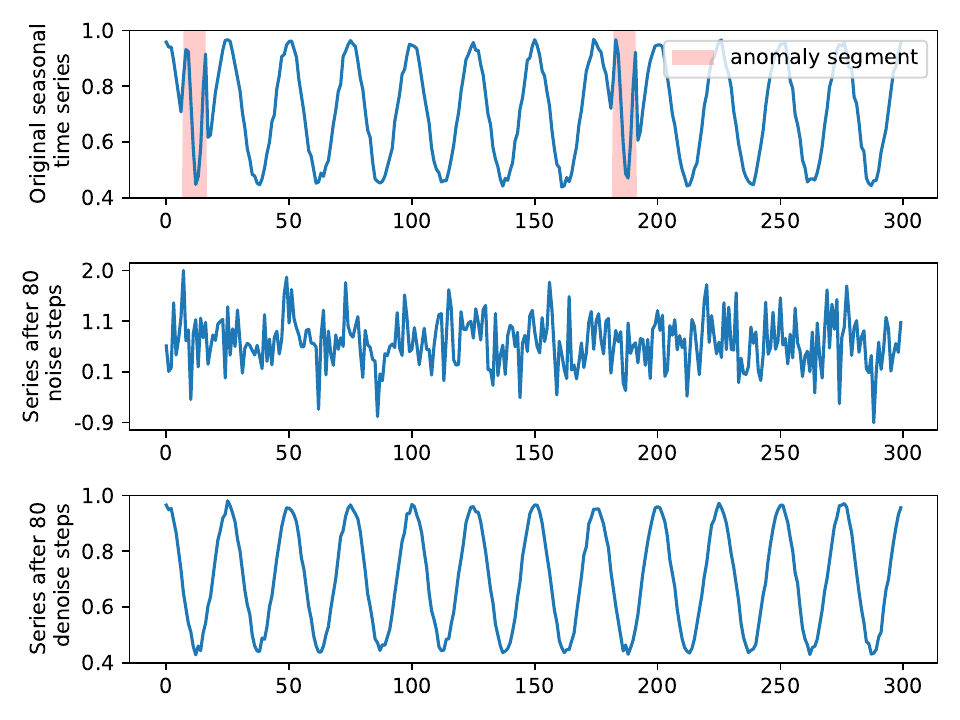}
    \caption{Top row: seasonal dataset window with two anomaly segments; middle row: data after 80 steps of Gaussian noise; bottom row: window denoised with the Diffusion model, where the anomaly segments are smoothed out, leading to larger reconstruction errors and improved AD performance.}
    \label{fig:denoise}
\end{figure}

We investigate whether diffusion models can be applied to AD in MTS. The fundamental idea, illustrated in Fig~\ref{fig:denoise}, is that the denoising process can smooth out the abnormal segments. This leads to increased differences between the original data and the reconstructions and therefore higher anomaly scores and better detection performance. Our input is a multivariate time series $\mathbf{X}^0 \in \mathbb{R}^{D\times T}$, where T is the sequence length and D is the number of features.  

We train two diffusion-based models: 
\vspace{-0.4em}
\begin{itemize}
    \item a diffusion model that learns to denoise a corrupted time series input:  $\mathbf{X}^M=noise(\mathbf{X}^0)$. 
    \item a diffusion model that learns to denoise an autoencoder reconstruction that was corrupted: $\hat{\mathbf{X}}^M=noise(\hat{\mathbf{X}}^0)$, where $\hat{\mathbf{X}}^0=autoencoder(\mathbf{X}^0)$.
\end{itemize}  

We jointly train the autoencoder and diffusion modules in the second model, which we refer to as \emph{DiffusionAE}. We use the distance between the original time series elements and the denoised elements as an anomaly score for both models.

Our contributions are summarized below:
\vspace{-0.4em}
\begin{enumerate}
    \item we apply diffusion models to AD in multivariate time series: we train two diffusion-based models, which outperform strong Transformer-based methods on synthetic datasets and are competitive on real world data

    \item we quantitatively show that the DiffusionAE model is more robust to different levels of outlier contamination and to the number of anomaly types
\end{enumerate}

Our results demonstrate the potential of diffusion models for anomaly detection in multivariate time series. We release the code at \url{https://github.com/fbrad/DiffusionAE}.

\section{Related Work}

\subsection{Classical methods}
While classical methods do not explicitly model the temporal information, they have been used for time series AD and they are competitive with deep methods on univariate time series~\cite{Rewicki2022}. 

The OC-SVM~\cite{Scholkopf2001} is a semi-supervised approach for AD in which the model is trained on normal data and predicts outliers if the points lie outside the learned boundary. It has been applied to synthetic and network intrusion datasets~\cite{Zhang2008,Zhang2007,Ma2003}. Local Outlier Factor (LOF~\cite{Breunig2000}) is an unsupervised method that detects anomaly points if they have a lower density than points in their neighbourhood. IsolationForest~\cite{Liu2008} fits binary decision trees to the data points and predicts anomalies if they require fewer splits. It can be adapted to time series AD by aggregating predictions over windows.

\subsection{Deep learning methods}
Most deep learning methods for AD in time series can be classified as \emph{forecasting-based} or \emph{reconstruction-based}. Forecasting models are usually trained autoregressively to predict the next step. The anomaly score is then given by how much the predicted values differ from the input. LSTM-NDT~\cite{Hundman2018} leverages an LSTM~\cite{Hochreiter1997} and an unsupervised dynamic thresholding approach. GDN~\cite{Deng2021} uses graph neural networks to learn dependencies between sensors and predicts anomalies by measuring the deviations from the graph structure.

Many reconstruction-based deep methods employ autoencoders, with different choices regarding the encoder/decoder architecture, stochasticity, cost function etc. Anomalous examples are scored based on the deviation of the reconstruction from the input. USAD~\cite{Audibert2020} trains autoencoders adversarially, while DAGMM~\cite{Zong2018} combines a deep autoencoder with a Gaussian Mixture Model to detect anomalies based on density estimation. MSCRED~\cite{Zhang2019} combines a fully convolutional architecture~\cite{Long2015} with an attention-based ConvLSTM~\cite{Shi2015} to extract relevant feature maps across timesteps at different scales. OmniAnomaly~\cite{Su2019} combines variational autoencoders~\cite{Kingma2013} with gated recurrent units~\cite{Cho2014} to map inputs to stochastic latent variables and to learn time dependencies in both latent and observation space. InterFusion \cite{Li2021} trains a hierarchical VAE with a two-view embeddings space to learn temporal-aware intermetric dependencies. To better exploit structure, graph-based approaches have also been used. MTAD-GAT~\cite{Zhao2020} uses two graph attention layers to learn dependencies between univariate features and temporal information.

Reconstruction-based methods also employ generative adversarial networks (GAN~\cite{Goodfellow2014}) to produce realistic temporal data. MAD-GAN~\cite{Li2019} trains a LSTM-based GAN and uses an anomaly score that combines the reconstruction error from the generator with the discriminator prediction.

Recently, Transformer-based~\cite{Vaswani2017} architectures have been succesfully applied to MTS. TranAD~\cite{Tuli2022} outperforms many deep and classical methods by using a two-step reconstruction approach in which the first reconstruction is used to improve the attention scores and representations of the second reconstruction, by amplifying the deviations from the input. AnomalyTransformer~\cite{Xu2022} exploits a property of the anomaly points, called the adjacent-concentration bias, which states that abnormal points are less likely to build strong associations with the entire time series. They improve the Transformer architecture via a two-branch attention mechanism, which consists of the classical self-attention module (to build associations over the entire series) and a Gaussian kernel (to build adjacent associations).

For an in-depth analysis of Transformer-based methods for AD  we refer the reader to surveys covering different time series tasks~\cite{Wen2022,Darban2022,Choi2021}. Several other surveys~\cite{Garcia2021,Pang2020,Braei2020} cover both classical and deep learning approaches, as well as anomaly detection methods for other modalities.

\subsection{Diffusion-based methods}
Denoising diffusion probabilistic models (DDPM~\cite{Ho2020,Dickstein2015}) have become a powerful class of generative methods, with successes across different modalities and tasks, such as image synthesis~\cite{Rombach2022,Dhariwal2021}, video generation~\cite{Ho2022}, text-to-sound generation~\cite{Yang2022} and controllable text generation~\cite{Li2022}.

Diffusion models have been used for AD on images \cite{Zhang2023,Wyatt2022}. The anomaly score is obtained by adding noise to an image then denoising it and measuring the difference between the original image and its reconstruction. DDPMs have also been used for time series forecasting~\cite{Rasul2021}. Concurrently to our work, ImDiffusion~\cite{chen2023imdiffusion} combines diffusion models and time series imputation to perform anomaly detection.

\section{Methods}

\begin{figure}[ht!]
    \centering
    \includegraphics[width=\linewidth]{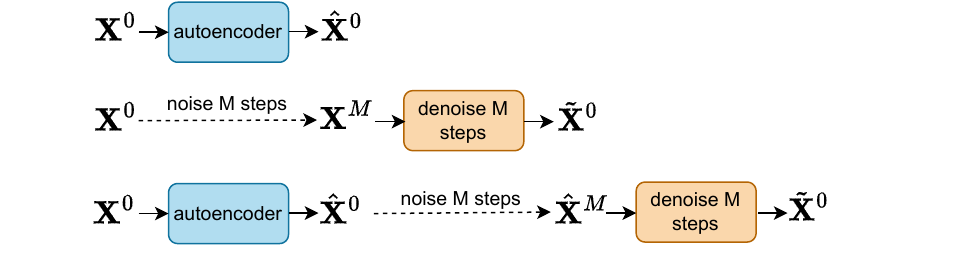}
    \caption{Test time anomaly detection flow for our models: the Transformer-based autoencoder, the Diffusion model and DiffusionAE, in which the reverse process is conditioned on the autoencoder reconstruction.}
    \label{fig:models}
\end{figure}
We train two diffusion-based models and an autoencoder baseline. We compare our methods to TranAD~\cite{Tuli2022} and AnomalyTransformer~\cite{Xu2022}, two state-of-the-art Transformer-based AD models on time series. All the models are reconstruction-based: reconstruction errors should be larger for anomalies than for normal samples. We illustrate the anomaly detection flow of our models in Figure~\ref{fig:models}.     

\subsection{Autoencoder Model}
We use a Transformer-based autoencoder $AE_{\phi}(.)$ as a baseline. The architecture is an encoder-decoder $AE_{\phi}(\mathbf{X}^0) = f_{\phi''}^{dec}(\mathbf{X}^0, f_{\phi'}^{enc}(\mathbf{X}^0))$, where $\phi=[\phi'; \phi'']$. It takes an input sequence $\mathbf{X}^0=[\mathbf{e}_1, \mathbf{e}_2, ..., \mathbf{e}_T]$, where each timestep element $\mathbf{e}_t \in \mathbb{R}^D$ is a feature vector, and outputs a reconstructed input sequence $\mathbf{\hat{X}^0}\in \mathbb{R}^{D\times T}$, where: $$\mathbf{\hat{X}^0} = AE_{\phi}(\mathbf{X}^0) = [\mathbf{e}_1', \mathbf{e}_2', ..., \mathbf{e}_T']$$
The anomaly score at timestep t is 
$$\textbf{s}_t=\norm{\mathbf{e}_t-\mathbf{e}_t'}^2,$$where $\mathbf{e}_t'=\hat{\mathbf{X}}^0_{[:,t]}$ is the reconstruction of the input $\mathbf{e}_t$.

Unlike recurrent-based autoencoders, Transformers have an advantage in reconstructing the input more efficiently. This is because the decoder can attend to the encoder states within a limited number of steps. However, this ease of reconstruction may lead to the model not learning a meaningful latent representation. To address this, we employ a modified Transformer architecture. In this modified version, we introduce a bottleneck~\cite{Wang2021,Montero2021}, by decoding from a single fixed representation
$$\mathbf{z} = \frac{1}{T}\sum_{i=1}^T \mathbf{h}_i,$$ where $[\mathbf{h}_1, \mathbf{h}_2, ..., \mathbf{h}_T] = f_{\phi'}^{enc}([\mathbf{e}_1, \mathbf{e}_2, ..., \mathbf{e}_T])$ are the outputs from the encoder. Thus, the decoder has direct access to the series representation $\mathbf{z}$ instead of the contextualized representation $\mathbf{h}_i$ of each timestep in the encoder. Specifically, the cross-attention in the decoder is
$$A(\mathbf{Q}, \mathbf{K}, \mathbf{V}) = softmax(\frac{\mathbf{Q}\mathbf{K}^T}{\sqrt{d}})\mathbf{V},$$ where $\mathbf{K}=\mathbf{V}=[\mathbf{z}] \in \mathbb{R}^{1\times d}$ are the key and value matrices containing $\mathbf{z}$, and $\mathbf{Q}=[\mathbf{s}_1, \mathbf{s}_2, ..., \mathbf{s}_T] \in \mathbb{R}^{T\times d}$ is the query matrix containing all the representations $\mathbf{s_i}$ from the first multi-head self-attention layer in the decoder.

\subsection{Diffusion Model}

\begin{algorithm}[t!]
  \caption{Training the Diffusion Model}
\begin{algorithmic}
  \STATE {\bfseries Input:} noise level $N \in \mathbb{N}^+$, data $\mathbf{X}^0 \sim q(\mathbf{X}^0), \mathbf{X}^0\in \mathbb{R}^{D\times T}$
  \REPEAT
  \STATE Initialize $n \sim \mathrm{Uniform}({1, \ldots, N})$ and $\epsilon \sim {\cal{N}}(\mathbf{0}, \mathbf{I})$
  \STATE Compute loss for the input $\mathbf{X}^n$ ($\mathbf{X}^0$ after n steps of noise):
  \STATE \;\;\;$L_{dif} = \| \mathbf{\epsilon} - \mathbf{\epsilon}_\theta (\sqrt{\bar{\alpha}_n} \mathbf{X}^0 + \sqrt{1 - \bar{\alpha}_n} \mathbf{\epsilon}, n ) \|^ 2$
  \STATE Take gradient step on $\nabla_\theta L_{dif}$
  \UNTIL{converged}
\end{algorithmic}
\label{algo1}
\end{algorithm}

\begin{algorithm}[t!]
  \caption{Anomaly Detection using the Diffusion Model}
\begin{algorithmic}
  \STATE {\bfseries Input:} noise level $M \in \mathbb{N}^+$, test data $\mathbf{X}^0 \in \mathbb{R}^{D\times T}$ 
  \STATE $\mathbf{X}^{M} = \sqrt{\bar{\alpha}_M} \mathbf{X}^{0} + \epsilon \sqrt{1 - \bar{\alpha}_M}$, where $\epsilon \in \mathcal{N}(\mathbf{0}, \mathbf{I})$
  \STATE $\tilde{\mathbf{X}}^{M}=\mathbf{X}^{M}$
  \FOR{n = M,\;\ldots, 1}
  \STATE $\mathbf{z} \sim {\cal{N}}(\mathbf{0}, \mathbf{I})$ if $n > 1$ else $\mathbf{z} = \mathbf{0}$
  \STATE $\tilde{\mathbf{X}}^{n-1} = \frac{1}{\sqrt{\alpha_n}}(\tilde{\mathbf{X}}^n-\frac{\beta_n}{\sqrt{1-\bar{\alpha}_n}}\epsilon_\theta(\tilde{\mathbf{X}}^n, n))+\tilde{\beta}_n\mathbf{z}$
  \ENDFOR
  \STATE $\mathbf{s}_t = \frac{1}{D} \| \mathbf{X}^0_{[:,t]} - \tilde{\mathbf{X}}^0_{[:,t]} \|^2$, for $t \in 1\ldots T$
  \STATE \textbf{Output: }anomaly scores $\mathbf{s} = [\mathbf{s}_1, \mathbf{s}_2, \ldots, \mathbf{s}_T]$\end{algorithmic}
\label{algo2}
\end{algorithm}

\begin{algorithm}[t!]
  \caption{Training the DiffusionAE Model}
\begin{algorithmic}
  \STATE {\bfseries Input:} noise level $N \in \mathbb{N}^+$, data $\mathbf{X}^0 \sim q(\mathbf{X}^0), \mathbf{X}^0\in \mathbb{R}^{D\times T}$, $\lambda \in \mathbb{R}^+$
  \REPEAT
  \STATE Get reconstruction from autoencoder: $\hat{\mathbf{X}}^0 = AE_\phi(\mathbf{X}^0)$
  \STATE Compute autoencoder loss: $L_{AE} = MSE(\hat{\mathbf{X}}^0, \mathbf{X}^0)$
  \STATE Initialize $n \sim \mathrm{Uniform}({1, \ldots, N})$ and $\epsilon \sim {\cal{N}}(\mathbf{0}, \mathbf{I})$
  \STATE Compute diffusion loss for the input $\hat{\mathbf{X}}^n$ ($\hat{\mathbf{X}}^0$ after n steps of noise):
  \STATE \;\;\;$L_{Dif} = \| \mathbf{\epsilon} - \mathbf{\epsilon}_\theta (\sqrt{\bar{\alpha}_n} \hat{\mathbf{X}}^0 + \sqrt{1 - \bar{\alpha}_n} \mathbf{\epsilon}, n ) \|^ 2$\;
  \STATE Take gradient step on $\nabla_{\phi,\theta} L$, where $L = L_{AE} + \lambda L_{Dif} $
  \UNTIL{converged}
\end{algorithmic}
\label{algo3}
\end{algorithm}

\begin{algorithm}[t!]
  \caption{Anomaly Detection using DiffusionAE}
\begin{algorithmic}
  \STATE {\bfseries Input:} noise level $M \in \mathbb{N}^+$, data $\mathbf{X}^0 \in \mathbb{R}^{D\times T}$
  \STATE Get reconstruction from autoencoder: $\hat{\mathbf{X}}^0 = AE_\phi(\mathbf{X}^0)$
  \STATE $\hat{\mathbf{X}}^M = \sqrt{\bar{\alpha}_M} \hat{\mathbf{X}}^0 + \epsilon \sqrt{(1 - \bar{\alpha}_M)}$, where $\epsilon \in \mathcal{N}(\mathbf{0}, \mathbf{I})$
  \STATE $\tilde{\mathbf{X}}^M=\hat{\mathbf{X}}^M$
  \FOR{n = M,\;\ldots, 1}
  \STATE $\mathbf{z} \sim {\cal{N}}(\mathbf{0}, \mathbf{I})$ if $n > 1$ else $\mathbf{z} = \mathbf{0}$
  \STATE $\tilde{\mathbf{X}}^{n-1} = \frac{1}{\sqrt{\alpha_n}}(\tilde{\mathbf{X}}^n-\frac{\beta_n}{\sqrt{1-\bar{\alpha}_n}}\epsilon_\theta(\tilde{\mathbf{X}}^n, n))+\tilde{\beta}_n\mathbf{z}$
  \ENDFOR
  \STATE $\mathbf{s}_t = \frac{1}{D} \| \mathbf{X}^0_{[:,t]} - \tilde{\mathbf{X}}^0_{[:,t]} \|^2$, for $t \in 1\ldots T$
  \STATE \textbf{Output: }anomaly scores $\mathbf{s} = [\mathbf{s}_1, \mathbf{s}_2, \ldots, \mathbf{s}_T]$
\end{algorithmic}
\label{algo4}
\end{algorithm}

Our model is based on a popular implementation for image synthesis\footnote{\url{https://huggingface.co/blog/annotated-diffusion}}. The denoising network $\epsilon_\theta$ is a U-Net~\cite{Ronneberger2015} architecture with classic Resnet~\cite{He2016} blocks and weight standardized convolutional layers~\cite{Qiao2019} instead of regular convolutions. 
The downsampling uses padding at each layer such that both the temporal and feature dimensions can be halved. The downsampling factors are [2, 4] for the synthetic datasets and [2, 4, 8] for the real-world datasets.

Our input $\mathbf{X}^0=[\mathbf{e}_1, \mathbf{e}_2, ..., \mathbf{e}_T] \in \mathbb{R}^{D\times T}$ is a multivariate time series, which can be seen as a single-channel image. Let $q(\mathbf{X}^0)$ be the distribution over the time series, from which we can sample $\mathbf{X}^0 \sim q(\mathbf{X}^0)$. During the forward process, we gradually add Gaussian noise on the previous input, starting from $\mathbf{X}^0$: $$q(\mathbf{X}^n | \mathbf{X}^{n-1}) =  \mathcal{N}(\mathbf{X}^{n}; \sqrt{1-{\beta}_n} \mathbf{X}^{n-1}, \beta_n \mathbf{I}),$$where $\beta_n\in(0, 1)$ is a fixed variance linearly increasing with $n$.

During the reverse process, we gradually remove noise from the input, starting from the corrupted time series. This process is modeled as 
$$p_\theta(\mathbf{X}^{n-1} | \mathbf{X}^{n}) = \mathcal{N}(\mathbf{X}^{n-1}; \mu_\theta(\mathbf{X}^n, n), \tilde{\beta}_n\mathbf{I}),$$ where $\tilde{\beta}_n=\frac{1-\bar{\alpha}_{n-1}}{1-\bar{\alpha}_n}\beta_n$ and is learned by a neural network, with $\alpha_{n}=1-\beta_{n}$ and $\bar{\alpha}_n=\prod_{s=1}^{n}\alpha_{s}$. Instead of learning to predict $\mu_\theta(\mathbf{X}^n, n)$, a more efficient approach~\cite{Ho2020} trains a network $\epsilon_\theta$ to predict the noise $\epsilon \sim \mathcal{N}(\mathbf{0}, \mathbf{I})$ given $\mathbf{X}^n$. The mean is then computed as 
$$\mu_\theta(\mathbf{X}^n, n) = \frac{1}{\sqrt{\alpha_n}}(\mathbf{X}^n-\frac{\beta_n}{\sqrt{1-\bar{\alpha}_n}}\epsilon_\theta(\mathbf{X}^n, n)).$$ Training the network involves minimizing the loss term 
$$L=\norm{\epsilon-\epsilon_\theta(\sqrt{\bar{\alpha}_n}\mathbf{X}^0+\sqrt{1-\bar{\alpha}_n}\epsilon, n)}^2$$ at different noise levels $n$ as can be seen in Algorithm~\ref{algo1}.

At test time, we add noise to an input $\mathbf{X}^0$ then denoise it back: $\mathbf{X}^0 \xrightarrow{noise} \mathbf{X}_{noisy} \xrightarrow{denoise} \tilde{\mathbf{X}}^0$. Specifically, we add Gaussian noise on the previous input M times, starting from $\mathbf{X}^0$. We can sample directly at noise level $M$: $\mathbf{X}^M \sim q(\mathbf{X}^n|\mathbf{X^0})$, where $$q(\mathbf{X}^n | \mathbf{X}^{0}) = \mathcal{N}(\mathbf{X}^{n}; \sqrt{\bar{\alpha}_n} \mathbf{X}^{0}, (1 - \bar{\alpha}_n) \mathbf{I}).$$ We then iteratively denoise the previous input $M$ times, starting from $\tilde{\mathbf{X}}^M=\mathbf{X}^M$. The distance between the reconstruction $\tilde{\mathbf{X}}^0$ and the original sample $\mathbf{X}^0$ is then used as an anomaly score. We test the anomaly detector using several noise levels $M\leq N$, where N is the noise level during training. The entire procedure is described in Algorithm~\ref{algo2} and briefly illustrated in Figure~\ref{fig:denoise}.

\subsection{DiffusionAE Model}

\begin{table}[t!]
    \setlength{\tabcolsep}{1.5pt} 
    \caption{Datasets statistics}
    \begin{center}
        \begin{tabular}{ lccccccc }
            \toprule
            \shortstack{Dataset} &
            \shortstack{Anomaly\\ type} &
            \shortstack{Train} &
            \shortstack{Val} &
            \shortstack{Test} &
            \shortstack{$\lvert D\rvert$} &
            \shortstack{Anomalies\\ in train} &
            \shortstack{Anomalies\\ in test} \\
            
            \midrule
            Global & point & 20,000 & 10,000 & 20,000 & 5 & \cmark & 6\% \\
            Contextual & point & 20,000 & 10,000 & 20,000 & 5 & \cmark & 6\% \\
            Seasonal & pattern & 20,000 & 10,000 & 20,000 & 5 & \cmark & 6\% \\
            Shapelet & pattern & 20,000 & 10,000 & 20,000 & 5 & \cmark & 6\% \\
            Trend & pattern & 20,000 & 10,000 & 20,000 & 5 & \cmark & 5\% \\ 
            \hdashline\noalign{\vskip 0.5ex}
            SWaT & pattern & 495,000 & 44,991 & 404,928 & 51 & \xmark & 12.38\% \\
            WADI & pattern & 1,048,571 & 30,000 & 142,801 & 123 & \xmark & 5.77\%  \\
            \bottomrule
        \end{tabular}
    \end{center}
    \label{tab: datasets}
\end{table}

This model is a jointly trained autoencoder $AE_{\phi}(.)$ and diffusion model $Dif_{\theta}(.)$. The autoencoder and diffusion architectures are the same as in the previous sections. During training, the autoencoder reconstructs the input time series: $\hat{\mathbf{X}}^0 = AE_{\phi}(\mathbf{X}^0)$. We then add noise on the reconstruction $\hat{\mathbf{X}}^0$. Distinctly from the diffusion-only model, the network $\epsilon_{\theta}$ learns to denoise the corrupted reconstruction of the input instead of the corrupted original sequence,  making the entire model trainable end-to-end.

We hypothesize that this joint training should help the diffusion be more robust to small noise levels in the data, since the autoencoder reconstruction can be seen as a perturbed input, onto which additional noise is added during the forward process. The training algorithm is described in Algorithm~\ref{algo3}.

At test time, we reconstruct the input $\mathbf{X}^0$ using the autoencoder. We then add noise to the reconstruction then denoise it back. 
We use the distance between the denoised input $\tilde{\mathbf{X}}^0$ and the original input $\mathbf{X}^0$ as the anomaly score. 
The AD procedure is described in Algorithm~\ref{algo4}.

\subsection{Baselines}

We compare our diffusion models to several Transformer-based methods: the previously described Autoencoder and two state-of-the-art models (TranAD~\cite{Tuli2022} and AnomalyTransformer~\cite{Xu2022}). We also evaluate a classical AD method (One-Class SVM model\cite{Scholkopf2001} with a linear kernel). 

For AnomalyTransformer, we apply early stopping based on the validation loss, while for TranAD and the Autoencoder, we select the model with the best validation $F1_K$-AUC. 

We keep the threshold selection criterion of each method. AnomalyTransformer uses the Gap Statistic method~\cite{Tibshirani2001} to determine the ratio $r \in (0, 1)$ of anomalous points in the validation dataset. It then sorts the test anomaly scores and selects a threshold $\delta$ such that a proportion $r$ of the scores are higher than $\delta$. TranAD uses the Peak Over Threshold~\cite{Siffer2017} algorithm to select the threshold based on the validation anomaly scores.  

We evaluate the $F1_K$-AUC metric based on the chosen threshold. We also evaluate the threshold-agnostic $ROC_K$-AUC for all the models.

\begin{figure}[t!]
    \centering
    \includegraphics[width=\linewidth]{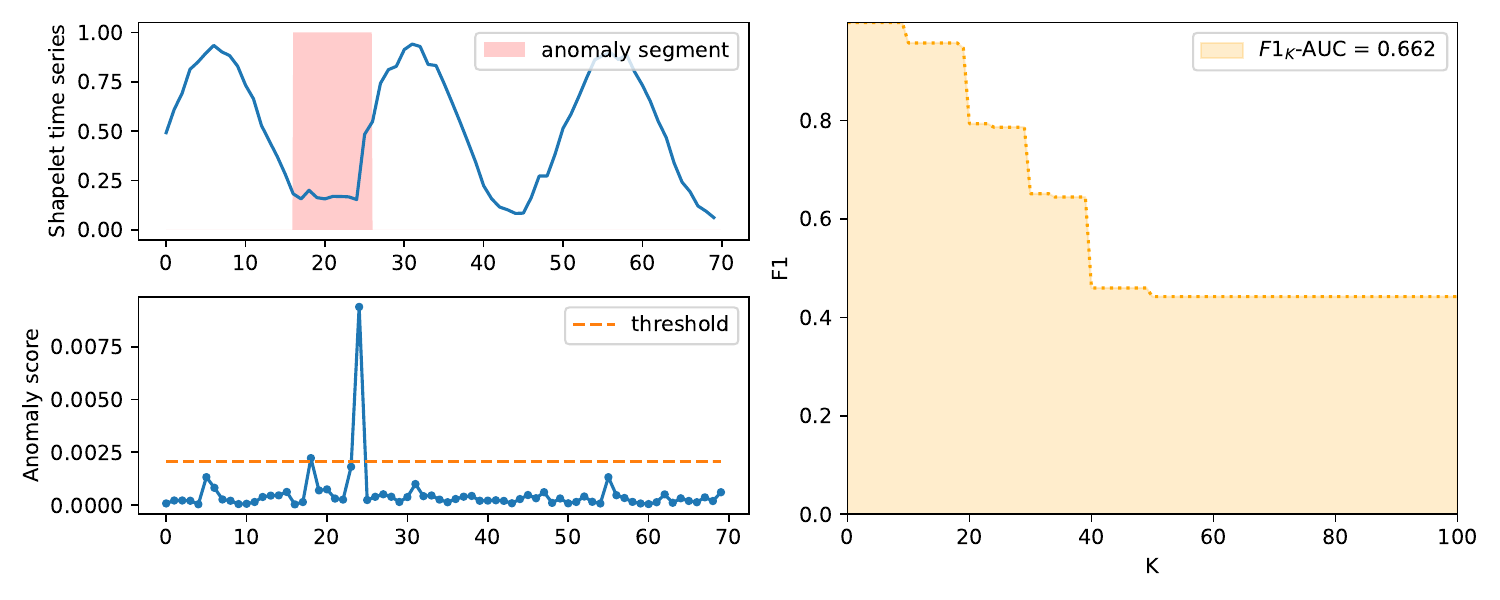}
    \caption{Left: subset of the Shapelet test set, with an anomaly segment (top). Based on the reconstruction score of this input (bottom), only 2 out of 11 points are detected as anomalies, but due to the point-adjustment protocol ($K=0$), all the points in the segment are counted as correct predictions. Right: $F1$-score for different levels of $K$ and a fixed threshold. Point-adjustment ($K=0$) leads to a high $F1$ score (0.88), but measuring the area under this curve (0.662) illustrates a more temperate performance.}
    \label{fig:F1K-AUC}
\end{figure}

\begin{figure}[t!]
    \centering
    \includegraphics[width=5cm]{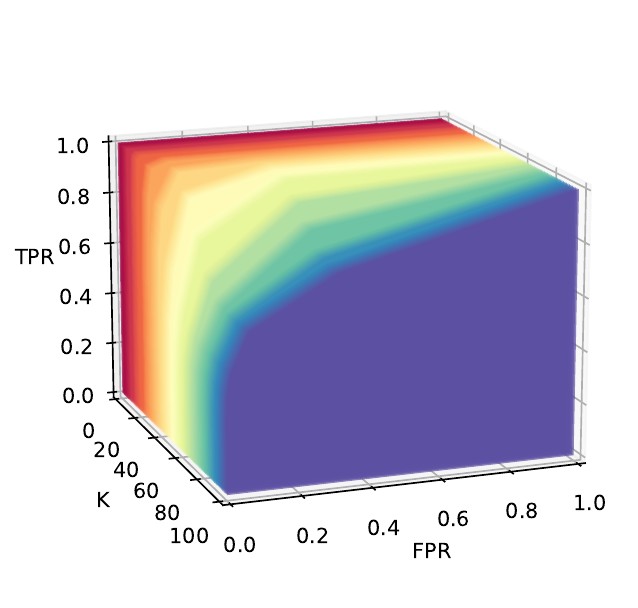}
    \caption{ROC curves for different $K$ values, computed on the Shapelet test set with a DiffusionAE model. $K=0$ is equivalent to the \emph{point-adjustment} protocol, which results in an overly optimistic AUROC = 1. Larger values of $K$ result in more realistic performances, with decreasing AUROCs. Our $ROC_K$-AUC metric is the area under a ROC curve computed over both thresholds and $K$ values.}
    \label{fig:rock}
\end{figure}

\subsection{Datasets}

\begingroup
\begin{table*}[t!]
    \setlength{\tabcolsep}{3.2pt} 
\caption{Model comparison. \textcolor{cb-green-sea}{Best results}. \textcolor{cb-burgundy}{Second best results.}}
    
    \begin{center}
        \begin{tabular}{ l cc|cc|cc|cc|cc|cc}
            \toprule
             &
            \multicolumn{2}{c}{DiffusionAE} &
            \multicolumn{2}{c}{Diffusion} &
            \multicolumn{2}{c}{Autoencoder} &
            \multicolumn{2}{c}{TranAD} &
            \multicolumn{2}{c}{AnomalyTransformer} &
            \multicolumn{2}{c}{OCSVM} 
            \\
            \midrule
            & \tiny{$F1_K$-AUC} & \tiny{$ROC_K$-AUC} & \tiny{$F1_K$-AUC} & \tiny{$ROC_K$-AUC} & \tiny{$F1_K$-AUC} & \tiny{$ROC_K$-AUC} & \tiny{$F1_K$-AUC} & \tiny{$ROC_K$-AUC} &
            \tiny{$F1_K$-AUC} & \tiny{$ROC_K$-AUC} &
            \tiny{$F1_K$-AUC} & \tiny{$ROC_K$-AUC} \\
            \midrule         
            Global & \textcolor{cb-green-sea}{\textbf{88.3$\pm$0.3}} & \textcolor{cb-green-sea}{\textbf{98.5$\pm$0.3}} & 86.1$\pm$2.0 & 96.3$\pm$0.8 & \textcolor{cb-burgundy}{\textbf{87.5$\pm$1.0}} & \textcolor{cb-burgundy}{\textbf{97.6$\pm$0.4}} & 10.4$\pm$0.2 & 51.0$\pm$0.5 & 12.6$\pm$5.2 & 51.9$\pm$1.5 & 10.8$\pm$0.0 & 9.2$\pm$0.0 \\
            Contextual & \textcolor{cb-green-sea}{\textbf{77.7$\pm$0.5}} & \textcolor{cb-green-sea}{\textbf{91.5$\pm$0.3}} & \textcolor{cb-burgundy}{\textbf{74.2$\pm$0.8}} & 88.8$\pm$0.7 & 70.0$\pm$1.4 & \textcolor{cb-burgundy}{\textbf{91.0$\pm$0.6}} & 10.0$\pm$0.6 & 50.6$\pm$1.1 & 8.3$\pm$0.8 & 50.4$\pm$0.3 & 10.9$\pm$0.0 & 29.3$\pm$0.0 \\
            Seasonal & \textcolor{cb-green-sea}{\textbf{94.6$\pm$0.4}} & \textcolor{cb-burgundy}{\textbf{99.6$\pm$0.1}} & \textcolor{cb-burgundy}{\textbf{93.8$\pm$0.6}} & \textcolor{cb-burgundy}{\textbf{99.6$\pm$0.1}} &  92.9$\pm$0.8 & \textcolor{cb-green-sea}{\textbf{99.7$\pm$0.1}} & 69.8$\pm$0.2 & 97.1$\pm$1.0 & 14.8$\pm$2.1 & 50.0$\pm$0.4 & 25.0$\pm$0.0 & 47.2$\pm$0.0 \\
            Shapelet & \textcolor{cb-green-sea}{\textbf{68.5$\pm$4.5}} &	\textcolor{cb-green-sea}{\textbf{92.8$\pm$1.1}} & \textcolor{cb-burgundy}{\textbf{67.3$\pm$2.5}} & 91.3$\pm$1.1 & 60.6$\pm$1.6 & \textcolor{cb-burgundy}{\textbf{92.3$\pm$1.0}} &  51.1$\pm$3.4 & 83.1$\pm$1.7 & 17.9$\pm$4.2 & 51.5$\pm$1.4 & 12.9$\pm$0.0 & 48.6$\pm$0.0 \\
            Trend & \textcolor{cb-burgundy}{\textbf{53.0$\pm$6.9}} & \textcolor{cb-burgundy}{\textbf{88.2$\pm$1.6}} & \textcolor{cb-green-sea}{\textbf{63.8$\pm$3.9}} & \textcolor{cb-green-sea}{\textbf{90.1$\pm$0.8}} & 42.7$\pm$22.5 & 86.3$\pm$2.7 & 33.1$\pm$15.7 & 83.4$\pm$0.8 & 16.3$\pm$3.6 & 53.0$\pm$0.8 & 8.9$\pm$0.0 & 32.3$\pm$0.0 \\ 
            \hdashline\noalign{\vskip 0.9ex}
            SWaT & \textcolor{cb-burgundy}{\textbf{54.0$\pm$0.5}} & 87.1$\pm$1.6 & 50.5$\pm$5.4 & 76.7$\pm$4.6 & \textcolor{cb-green-sea}{\textbf{56.7$\pm$0.7}} & \textcolor{cb-burgundy}{\textbf{89.2$\pm$1.4}} & 38.3$\pm$5.1 & \textcolor{cb-green-sea}{\textbf{89.5$\pm$0.7}} & 22.0$\pm$0.0 & 50.2$\pm$0.2 & 10.0$\pm$0.0 & 13.9$\pm$0.0 \\ 
            WADI & 12.4$\pm$0.4 & \textcolor{cb-burgundy}{\textbf{77.0$\pm$2.6}} & \textcolor{cb-burgundy}{\textbf{16.9$\pm$1.9}} & 76.5$\pm$3.2 & \textcolor{cb-green-sea}{\textbf{17.2$\pm$5.4}} & \textcolor{cb-green-sea}{\textbf{ 84.2$\pm$2.1}} & 11.2$\pm$0.4 & 55.1$\pm$14.5 & 10.9$\pm$0.0 & 50.0$\pm$0.1 & 3.9$\pm$0.0 & 31.1$\pm$0.0\\
            \bottomrule
        \end{tabular}
    \end{center}

    \label{tab: results}
\end{table*}
\endgroup

We evaluate all the models on both synthetic and real-world datasets. Detailed information regarding these datasets can be found in Table~\ref{tab: datasets}. We create five multivariate synthetic datasets. Each dataset has a different anomaly type based on the NeurIPS-TS~\cite{Lai2021} behaviour-driven taxonomy: point anomalies (global and contextual) and pattern anomalies (seasonal, shapelet and trend). Each generated synthetic dataset has 50,000 timesteps with 4 normal dimensions and one dimension containing anomalies. For the synthetic datasets, we ensure that the training, validation and test splits have a similar anomaly ratio.

We also evaluate the models on SWaT~\cite{Mathur2016} and WADI~\cite{Ahmed2017}, two popular datasets collected from real-world water systems sensors. Distinctly from the synthetic datasets, SWaT and WADI do not have anomalies in the training data and so we keep the original training split. We split their original test data into validation and test sets. The validation set of WADI has a similar anomaly ratio (5.79\%) to the test set (5.77\%). Due to a large anomaly segment in the raw SWaT test set, the resulting validation set has a slightly lower anomaly ratio than the test set (9.93\% vs 12.38\%).

\subsection{Preprocessing and training}
Let $\mathbf{\mathcal{T}}=[\mathbf{e}_1, \mathbf{e}_2, \ldots, \mathbf{e}_L] \in \mathbb{R}^{D\times L}$ be a multivariate time series with $L$ timesteps and $D$ dimensions. We split the time series $\mathbf{\mathcal{T}}$ into training, validation and test sets. Each subset is then split into equally-sized, non-overlapping windows of length $T$: $$\mathbf{W}_1=[\mathbf{e}_1, \mathbf{e}_2, \ldots, \mathbf{e}_{T}],$$ $$\mathbf{W}_2=[\mathbf{e}_{T+1}, \ldots, \mathbf{e}_{2T}],$$ $$\ldots$$ $$\mathbf{W}_{l}=[\mathbf{e}_{(l-1)T+1},\ldots, \mathbf{e}_{lT}],$$ where $l=\lfloor L/T \rfloor$ is the total number of windows. The training and detection procedures described in Algorithms 1-4 process a single window. To obtain anomaly scores for the entire series $\mathbf{\mathcal{T}}$, we concatenate the scores from all its windows.

We set $T=100$ for the Autoencoder and the diffusion-based models. For TranAD we keep $T=10$ for all the datasets, while for AnomalyTransformer we set $T=10$ for the synthetic data and $T=100$ for the real-world datasets.

We normalize the data using maximum absolute scaling for the Global, Contextual, Seasonal and Shapelet datasets and min-max scaling for the other datasets. For Trend, SWaT and WADI, we fit the scaler to the training data, normalizing each feature across the timesteps. We then rescale the validation and test sets and clip the values to the interval $[0, 1]$.

We search for the best hyperparameters across batch size $b \in \{8, 16, 32, 64, 128\}$, learning rate $lr \in \{1e-3,1e-4\}$ and diffusion loss weight term $\lambda \in \{0.1, 0.01\}$.  We train the diffusion-based models with noise level $N=100$ and test them with several noise levels $M \in \{10, 20, 50, 60, 80\}$. 

To ensure that the diffusion process is not conditioned on excessively noisy reconstructions in the DiffusionAE model, we train the autoencoder independently for 5 epochs, and then proceed to jointly train both components.

\subsection{Evaluation}

Time series AD is often evaluated using a protocol called \emph{point-adjustment}~\cite{Xu2018,Su2019,Shen2020}. Given an anomalous segment, it considers all point predictions in that segment to be correct if at least one of them is correct. This strategy can lead to an overestimation of the true capability of an AD system, since two different methods can have equal F1-scores, even if one correctly predicts all the points in an abnormal segment and the other one only guesses one point. Therefore, \emph{point-adjustment} results in large F1-scores across different datasets.

Due to this overestimation, a more rigurous evaluation protocol in time series AD has been called for~\cite{Kim2022,Sorbo2023}. An alternative protocol called $PA\%K$~\cite{Kim2022} computes point-adjustment at different $K$ values. Specifically, all predictions in an outlier segment are considered anomalous if at least $K\%$ of the points in the segment have been correctly predicted. Setting $K=0$ is equivalent to point-adjustment and leads to overestimation of the performance, while $K=100$ penalizes the model for even one point mistake. 

\begin{figure*}[t!]
    \centering
    \includegraphics[width=\linewidth]{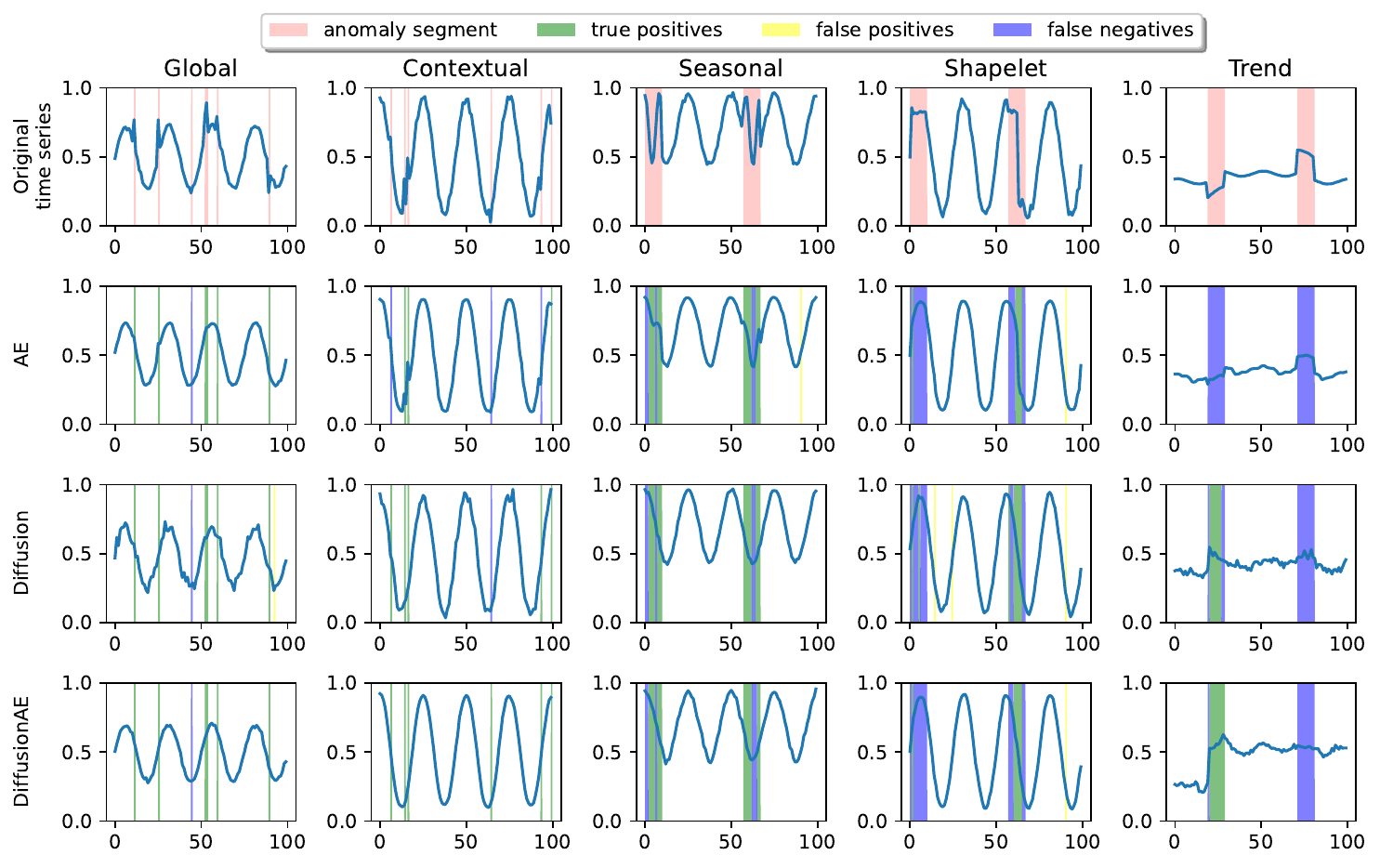}
    \caption{Reconstructions on the synthetic datasets for all the models. }
    \label{fig:reconstructions}
\end{figure*}

By measuring the F1-score at different levels of $K$, we can compute an area under this curve, which we call $F1_K$-AUC. We illustrate the necessity for this metric in Figure~\ref{fig:F1K-AUC}. 

We search for 50 evenly-spaced threshold values 
$$\delta \in \left\{\frac{k \textbf{s}_{max}}{50} \mid k = 0, 1, 2, \ldots, 49\right\},$$
where $\textbf{s}_{max} = \max_{t=1,L} \textbf{s}_t$ is the highest anomaly score in the sequence. We pick the threshold $\delta$ that results in the highest $F1_K$-AUC on the validation set for each model and then use it to for the test sets evaluation.

While $F1_K$-AUC removes the dependency on $K$, it is still evaluated for a given threshold $\delta$. To remove this dependency as well, we compute a modified $ROC$ curve, by measuring the true positive rates and false positive rates across thresholds and $K$ values, where $K \in \{0, 1, 2,\ldots 100\}$. We refer to the area under this ROC curve as $ROC_K$-AUC. The $ROC_K$-AUC is thus useful to compare AD algorithms in MTS regardless of the threshold selection. For easier visualization, we illustrate the ROC curves for each $K$ in Figure~\ref{fig:rock}, where the AUCs decrease for larger $K$ ratios.

\section{Results}
The results for all the models are shown in Table~\ref{tab: results}, with standard deviation computed over 5 runs. Diffusion-based models outperform the other models on all the synthetic datasets based on $F1_K$-AUC and $ROC_K$-AUC. DiffusionAE obtains higher $F1_K$-AUC scores than the other models on 4 out of 5 synthetic datasets. 
On the Trend dataset, the Diffusion model is better than the DiffusionAE model in terms of $F1_K$-AUC, but the gap closes when comparing the threshold-agnostic $ROC_K$-AUCs. 

DiffusionAE is also overall better on the real-world datasets than the Diffusion model, suggesting that the diffusion process benefits from conditioning on the autoencoder reconstruction. While both models obtain higher $F1_K$-AUC than the OCSVM and the two strong neural models (TranAD and AnomalyTransformer), the Transformer-based autoencoder is surprisingly effective, surpassing them by a small margin. However, the diffusion-based models rank second best overall on real world sets.

We notice that the best models on 5 out of 7 datasets have both the highest $F1_K$-AUC and $ROC_K$-AUC, respectively. To assess the relationship between the two metrics, we measure the Spearman's rank correlation coefficient, which is 0.89, indicating a strong correlation. This supports $ROC_K$-AUC as an additional metric for model selection in time series AD.

\begin{figure*}[t!]
    \centering
    \includegraphics[width=\linewidth]{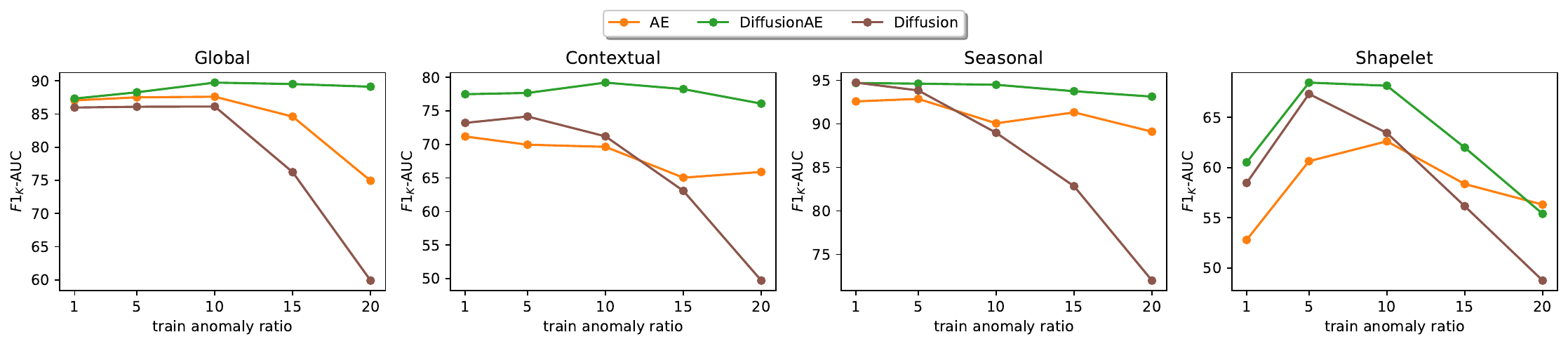}
    \caption{$F1_K$-AUC for different ratios of anomaly in the training data. All the models are evaluated on the same test split with 5\% anomalies}
    \label{fig:ratios}
\end{figure*}

\subsection{Qualitative analysis}
We illustrate the reconstruction of windows from each synthetic dataset for the autoencoder and the diffusion-based models in Figure~\ref{fig:reconstructions}. The autoencoder reconstructions of anomalous segments are generally more similar to the original sequence. This leads to smaller reconstruction error and therefore more false negatives, such as the first anomaly in the Contextual set or the anomaly segments in the Trend set.

We observe the Diffusion model smooths out the anomaly segments on all the datasets. However, the denoising process is not perfect, as can be seen in the Global and Contextual sets, where the reconstruction is more jittery. 

The DiffusionAE model is conditioned on the autoencoder's reconstruction, which has already smoothed out some of the abnormal segments. The denoising process based on this reconstruction can further smooth out the remaining irregularities, as can be seen in the Contextual set. By chaining two reconstructions, the DiffusionAE model can even out the outlier fragments more effectively, resulting in increased reconstruction errors and a higher number of true positives.

\subsection{Anomaly ratio analysis}
We evaluate our models' performance on the synthetic datasets for different outlier contamination levels, by varying the anomaly ratios in the training set. We compare them to the Transformer-based autoencoder, which was the best performing baseline on the synthetic sets. 

We regenerate the train and validation splits with r\% anomalies, where $r \in \{1, 5, 10, 15, 20\}$. For each ratio $r$, we select the best model based on its validation $F1_K$-AUC score. We then evaluate all the models on the same test split, which has 5\% outliers.

\begin{figure}[t!]
    \centering
    \includegraphics[width=\linewidth]{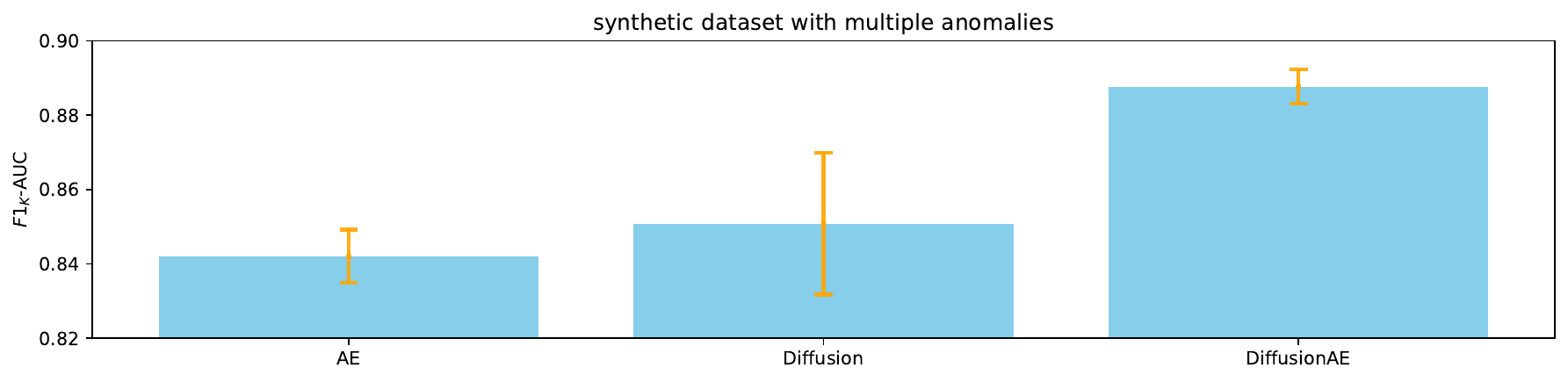}
    \caption{$F1_K$-AUC on the synthetic dataset containing 4 anomalous dimensions out of 5. The error bars are plotted with standard deviations over 5 runs. }
    \label{fig:multianomaly}
\end{figure}

We notice in Figure~\ref{fig:ratios} that the autoencoder and the Diffusion model are less robust to higher anomaly ratios in the training set. The Diffusion model exhibits the most significant drop in $F1_K$-AUC across all the datasets for more than 10\% anomalies in the training set. 

The performance of DiffusionAE does not degrade as much with more training outliers. For the Shapelet dataset, both diffusion-based models perform similarly for low anomaly ratios (1\% and 5\%), but the Diffusion model's performance more significantly drops for higher outlier contamination ($\geq$10\%). The relatively stable performance of the DiffusionAE model with respect to different anomaly ratios suggests that it may be better suited on real world scenarios in which the anomaly ratio changes drastically on new data.

\subsection{Multiple anomalies analysis}
All the synthetic datasets have 5 dimensions, of which only one is anomalous. We test our models on the opposite scenario, in which 4 dimensions have anomalies and only one dimension is normal. To increase the challenge of the dataset, each of the 4 dimensions contains a different type of anomaly: global, contextual, seasonal, and shapelet. Thus, within a given time interval, multiple types of anomaly segments may overlap. We set the anomaly ratio on each dimension to 1.25\%.

Results are plotted in Figure~\ref{fig:multianomaly}. Based on the $F1_K$-AUC score, both diffusion-based models outperform the autoencoder. Moreover, the DiffusionAE model outperforms the Diffusion model  by 4 $F1_K$-AUC points, while also having a smaller variance than both the other models. This suggests that the DiffusionAE model is more robust to the number of anomaly types in the dataset and may be more suitable for realistic scenarios where anomalous events produce abnormal values on several dimensions at once.

\section{Conclusions}
In this paper we introduce two diffusion-based models for anomaly detection in multivariate time series. The models outperform the classical and deep methods on all the synthetic sets and perform strongly on the real world sets. Upon qualitative inspection, the diffusion process smooths out the anomaly segments, leading to higher reconstruction errors and improved detection. 

We also introduce $ROC_K$-AUC, an extension of the PA\%K protocol to $ROC$-AUC, and illustrate its potential as an additional metric for model selection.

By conditioning the diffusion on an autoencoder reconstruction instead of the original data, the resulting DiffusionAE model leads to better overall performance on both dataset types. The DiffusionAE model is also more robust to different training anomaly ratios and can better handle multiple anomaly types in a set, suggesting that diffusion-based anomaly detection in time series benefits from this two-step approach.

\bibliographystyle{splncs04}
\bibliography{bibliography}

\begin{thebibliography}{10}
\providecommand{\url}[1]{\texttt{#1}}
\providecommand{\urlprefix}{URL }
\providecommand{\doi}[1]{https://doi.org/#1}

\bibitem{Ahmed2017}
Ahmed, C.M., Palleti, V.R., Mathur, A.P.: {WADI:} a water distribution testbed
  for research in the design of secure cyber physical systems. In: Tsakalides,
  P., Beferull{-}Lozano, B. (eds.) Proceedings of the 3rd International
  Workshop on Cyber-Physical Systems for Smart Water Networks, CySWATER@CPSWeek
  2017, Pittsburgh, Pennsylvania, USA, April 21, 2017. pp. 25--28. {ACM} (2017)

\bibitem{Audibert2020}
Audibert, J., Michiardi, P., Guyard, F., Marti, S., Zuluaga, M.A.: {USAD:}
  unsupervised anomaly detection on multivariate time series. In: Gupta, R.,
  Liu, Y., Tang, J., Prakash, B.A. (eds.) {KDD} '20: The 26th {ACM} {SIGKDD}
  Conference on Knowledge Discovery and Data Mining, Virtual Event, CA, USA,
  August 23-27, 2020. pp. 3395--3404. {ACM} (2020)

\bibitem{Garcia2021}
Bl{\'{a}}zquez{-}Garc{\'{\i}}a, A., Conde, A., Mori, U., Lozano, J.A.: A review
  on outlier/anomaly detection in time series data. {ACM} Comput. Surv.
  \textbf{54}(3),  56:1--56:33 (2022)

\bibitem{Braei2020}
Braei, M., Wagner, S.: Anomaly detection in univariate time-series: {A} survey
  on the state-of-the-art. CoRR  \textbf{abs/2004.00433} (2020)

\bibitem{Breunig2000}
Breunig, M.M., Kriegel, H., Ng, R.T., Sander, J.: {LOF:} identifying
  density-based local outliers. In: Chen, W., Naughton, J.F., Bernstein, P.A.
  (eds.) Proceedings of the 2000 {ACM} {SIGMOD} International Conference on
  Management of Data, May 16-18, 2000, Dallas, Texas, {USA}. pp. 93--104. {ACM}
  (2000)

\bibitem{chen2023imdiffusion}
Chen, Y., Zhang, C., Ma, M., Liu, Y., Ding, R., Li, B., He, S., Rajmohan, S.,
  Lin, Q., Zhang, D.: Imdiffusion: Imputed diffusion models for multivariate
  time series anomaly detection. arXiv preprint arXiv:2307.00754  (2023)

\bibitem{Cho2014}
Cho, K., van Merrienboer, B., G{\"{u}}l{\c{c}}ehre, {\c{C}}., Bahdanau, D.,
  Bougares, F., Schwenk, H., Bengio, Y.: Learning phrase representations using
  {RNN} encoder-decoder for statistical machine translation. In: Moschitti, A.,
  Pang, B., Daelemans, W. (eds.) Proceedings of the 2014 Conference on
  Empirical Methods in Natural Language Processing, {EMNLP} 2014, October
  25-29, 2014, Doha, Qatar, {A} meeting of SIGDAT, a Special Interest Group of
  the {ACL}. pp. 1724--1734. {ACL} (2014)

\bibitem{Choi2021}
Choi, K., Yi, J., Park, C., Yoon, S.: Deep learning for anomaly detection in
  time-series data: Review, analysis, and guidelines. {IEEE} Access
  \textbf{9},  120043--120065 (2021)

\bibitem{Darban2022}
Darban, Z.Z., Webb, G.I., Pan, S., Aggarwal, C.C., Salehi, M.: Deep learning
  for time series anomaly detection: {A} survey. CoRR  \textbf{abs/2211.05244}
  (2022)

\bibitem{Deng2021}
Deng, A., Hooi, B.: Graph neural network-based anomaly detection in
  multivariate time series. In: Thirty-Fifth {AAAI} Conference on Artificial
  Intelligence, {AAAI} 2021, Thirty-Third Conference on Innovative Applications
  of Artificial Intelligence, {IAAI} 2021, The Eleventh Symposium on
  Educational Advances in Artificial Intelligence, {EAAI} 2021, Virtual Event,
  February 2-9, 2021. pp. 4027--4035. {AAAI} Press (2021)

\bibitem{Dhariwal2021}
Dhariwal, P., Nichol, A.Q.: Diffusion models beat gans on image synthesis. In:
  Ranzato, M., Beygelzimer, A., Dauphin, Y.N., Liang, P., Vaughan, J.W. (eds.)
  Advances in Neural Information Processing Systems 34: Annual Conference on
  Neural Information Processing Systems 2021, NeurIPS 2021, December 6-14,
  2021, virtual. pp. 8780--8794 (2021)

\bibitem{Goodfellow2014}
Goodfellow, I.J., Pouget{-}Abadie, J., Mirza, M., Xu, B., Warde{-}Farley, D.,
  Ozair, S., Courville, A.C., Bengio, Y.: Generative adversarial networks. CoRR
   \textbf{abs/1406.2661} (2014)

\bibitem{He2016}
He, K., Zhang, X., Ren, S., Sun, J.: Deep residual learning for image
  recognition. In: 2016 {IEEE} Conference on Computer Vision and Pattern
  Recognition, {CVPR} 2016, Las Vegas, NV, USA, June 27-30, 2016. pp. 770--778.
  {IEEE} Computer Society (2016)

\bibitem{Ho2020}
Ho, J., Jain, A., Abbeel, P.: Denoising diffusion probabilistic models. In:
  Larochelle, H., Ranzato, M., Hadsell, R., Balcan, M., Lin, H. (eds.) Advances
  in Neural Information Processing Systems 33: Annual Conference on Neural
  Information Processing Systems 2020, NeurIPS 2020, December 6-12, 2020,
  virtual (2020)

\bibitem{Ho2022}
Ho, J., Salimans, T., Gritsenko, A.A., Chan, W., Norouzi, M., Fleet, D.J.:
  Video diffusion models. CoRR  \textbf{abs/2204.03458} (2022)

\bibitem{Hochreiter1997}
Hochreiter, S., Schmidhuber, J.: Long short-term memory. Neural Comput.
  \textbf{9}(8),  1735--1780 (1997)

\bibitem{Hundman2018}
Hundman, K., Constantinou, V., Laporte, C., Colwell, I.,
  S{\"{o}}derstr{\"{o}}m, T.: Detecting spacecraft anomalies using lstms and
  nonparametric dynamic thresholding. In: Guo, Y., Farooq, F. (eds.)
  Proceedings of the 24th {ACM} {SIGKDD} International Conference on Knowledge
  Discovery {\&} Data Mining, {KDD} 2018, London, UK, August 19-23, 2018. pp.
  387--395. {ACM} (2018)

\bibitem{Kim2022}
Kim, S., Choi, K., Choi, H., Lee, B., Yoon, S.: Towards a rigorous evaluation
  of time-series anomaly detection. In: Thirty-Sixth {AAAI} Conference on
  Artificial Intelligence, {AAAI} 2022, Thirty-Fourth Conference on Innovative
  Applications of Artificial Intelligence, {IAAI} 2022, The Twelveth Symposium
  on Educational Advances in Artificial Intelligence, {EAAI} 2022 Virtual
  Event, February 22 - March 1, 2022. pp. 7194--7201. {AAAI} Press (2022)

\bibitem{Kingma2013}
Kingma, D.P., Welling, M.: Auto-encoding variational bayes. In: Bengio, Y.,
  LeCun, Y. (eds.) 2nd International Conference on Learning Representations,
  {ICLR} 2014, Banff, AB, Canada, April 14-16, 2014, Conference Track
  Proceedings (2014)

\bibitem{Lai2021}
Lai, K., Zha, D., Xu, J., Zhao, Y., Wang, G., Hu, X.: Revisiting time series
  outlier detection: Definitions and benchmarks. In: Vanschoren, J., Yeung, S.
  (eds.) Proceedings of the Neural Information Processing Systems Track on
  Datasets and Benchmarks 1, NeurIPS Datasets and Benchmarks 2021, December
  2021, virtual (2021)

\bibitem{Li2019}
Li, D., Chen, D., Jin, B., Shi, L., Goh, J., Ng, S.: {MAD-GAN:} multivariate
  anomaly detection for time series data with generative adversarial networks.
  In: Tetko, I.V., Kurkov{\'{a}}, V., Karpov, P., Theis, F.J. (eds.) Artificial
  Neural Networks and Machine Learning - {ICANN} 2019: Text and Time Series -
  28th International Conference on Artificial Neural Networks, Munich, Germany,
  September 17-19, 2019, Proceedings, Part {IV}. Lecture Notes in Computer
  Science, vol. 11730, pp. 703--716. Springer (2019)

\bibitem{Li2022}
Li, X.L., Thickstun, J., Gulrajani, I., Liang, P., Hashimoto, T.B.:
  Diffusion-lm improves controllable text generation. CoRR
  \textbf{abs/2205.14217} (2022)

\bibitem{Li2023}
Li, Y., Peng, X., Zhang, J., Li, Z., Wen, M.: {DCT-GAN:} dilated convolutional
  transformer-based {GAN} for time series anomaly detection. {IEEE} Trans.
  Knowl. Data Eng.  \textbf{35}(4),  3632--3644 (2023)

\bibitem{Li2021}
Li, Z., Zhao, Y., Han, J., Su, Y., Jiao, R., Wen, X., Pei, D.: Multivariate
  time series anomaly detection and interpretation using hierarchical
  inter-metric and temporal embedding. In: Zhu, F., Ooi, B.C., Miao, C. (eds.)
  {KDD} '21: The 27th {ACM} {SIGKDD} Conference on Knowledge Discovery and Data
  Mining, Virtual Event, Singapore, August 14-18, 2021. pp. 3220--3230. {ACM}
  (2021)

\bibitem{Liu2008}
Liu, F.T., Ting, K.M., Zhou, Z.: Isolation forest. In: Proceedings of the 8th
  {IEEE} International Conference on Data Mining {(ICDM} 2008), December 15-19,
  2008, Pisa, Italy. pp. 413--422. {IEEE} Computer Society (2008)

\bibitem{Long2015}
Long, J., Shelhamer, E., Darrell, T.: Fully convolutional networks for semantic
  segmentation. In: {IEEE} Conference on Computer Vision and Pattern
  Recognition, {CVPR} 2015, Boston, MA, USA, June 7-12, 2015. pp. 3431--3440.
  {IEEE} Computer Society (2015)

\bibitem{Ma2003}
Ma, J., Perkins, S.: Time-series novelty detection using one-class support
  vector machines. In: Proceedings of the International Joint Conference on
  Neural Networks, 2003. vol.~3, pp. 1741--1745 vol.3 (2003)

\bibitem{Mathur2016}
Mathur, A.P., Tippenhauer, N.O.: Swat: a water treatment testbed for research
  and training on {ICS} security. In: 2016 International Workshop on
  Cyber-physical Systems for Smart Water Networks, CySWater@CPSWeek 2016,
  Vienna, Austria, April 11, 2016. pp. 31--36. {IEEE} Computer Society (2016)

\bibitem{Montero2021}
Montero, I., Pappas, N., Smith, N.A.: Sentence bottleneck autoencoders from
  transformer language models. In: Moens, M., Huang, X., Specia, L., Yih, S.W.
  (eds.) Proceedings of the 2021 Conference on Empirical Methods in Natural
  Language Processing, {EMNLP} 2021, Virtual Event / Punta Cana, Dominican
  Republic, 7-11 November, 2021. pp. 1822--1831. Association for Computational
  Linguistics (2021)

\bibitem{Pang2020}
Pang, G., Shen, C., Cao, L., van~den Hengel, A.: Deep learning for anomaly
  detection: {A} review. {ACM} Comput. Surv.  \textbf{54}(2),  38:1--38:38
  (2022)

\bibitem{Qiao2019}
Qiao, S., Wang, H., Liu, C., Shen, W., Yuille, A.: Micro-batch training with
  batch-channel normalization and weight standardization. arXiv preprint
  arXiv:1903.10520  (2019)

\bibitem{Rasul2021}
Rasul, K., Seward, C., Schuster, I., Vollgraf, R.: Autoregressive denoising
  diffusion models for multivariate probabilistic time series forecasting. In:
  Meila, M., Zhang, T. (eds.) Proceedings of the 38th International Conference
  on Machine Learning, {ICML} 2021, 18-24 July 2021, Virtual Event. Proceedings
  of Machine Learning Research, vol.~139, pp. 8857--8868. {PMLR} (2021)

\bibitem{Rewicki2022}
Rewicki, F., Denzler, J., Niebling, J.: Is it worth it? an experimental
  comparison of six deep- and classical machine learning methods for
  unsupervised anomaly detection in time series. CoRR  \textbf{abs/2212.11080}
  (2022)

\bibitem{Rombach2022}
Rombach, R., Blattmann, A., Lorenz, D., Esser, P., Ommer, B.: High-resolution
  image synthesis with latent diffusion models. In: {IEEE/CVF} Conference on
  Computer Vision and Pattern Recognition, {CVPR} 2022, New Orleans, LA, USA,
  June 18-24, 2022. pp. 10674--10685. {IEEE} (2022)

\bibitem{Ronneberger2015}
Ronneberger, O., Fischer, P., Brox, T.: U-net: Convolutional networks for
  biomedical image segmentation. In: Navab, N., Hornegger, J., III, W.M.W.,
  Frangi, A.F. (eds.) Medical Image Computing and Computer-Assisted
  Intervention - {MICCAI} 2015 - 18th International Conference Munich, Germany,
  October 5 - 9, 2015, Proceedings, Part {III}. Lecture Notes in Computer
  Science, vol.~9351, pp. 234--241. Springer (2015)

\bibitem{Scholkopf2001}
Sch{\"{o}}lkopf, B., Platt, J.C., Shawe{-}Taylor, J., Smola, A.J., Williamson,
  R.C.: Estimating the support of a high-dimensional distribution. Neural
  Comput.  \textbf{13}(7),  1443--1471 (2001)

\bibitem{Shen2020}
Shen, L., Li, Z., Kwok, J.T.: Timeseries anomaly detection using temporal
  hierarchical one-class network. In: Larochelle, H., Ranzato, M., Hadsell, R.,
  Balcan, M., Lin, H. (eds.) Advances in Neural Information Processing Systems
  33: Annual Conference on Neural Information Processing Systems 2020, NeurIPS
  2020, December 6-12, 2020, virtual (2020)

\bibitem{Shi2015}
Shi, X., Chen, Z., Wang, H., Yeung, D., Wong, W., Woo, W.: Convolutional {LSTM}
  network: {A} machine learning approach for precipitation nowcasting. In:
  Cortes, C., Lawrence, N.D., Lee, D.D., Sugiyama, M., Garnett, R. (eds.)
  Advances in Neural Information Processing Systems 28: Annual Conference on
  Neural Information Processing Systems 2015, December 7-12, 2015, Montreal,
  Quebec, Canada. pp. 802--810 (2015)

\bibitem{Siffer2017}
Siffer, A., Fouque, P., Termier, A., Largou{\"{e}}t, C.: Anomaly detection in
  streams with extreme value theory. In: Proceedings of the 23rd {ACM} {SIGKDD}
  International Conference on Knowledge Discovery and Data Mining, Halifax, NS,
  Canada, August 13 - 17, 2017. pp. 1067--1075. {ACM} (2017)

\bibitem{Dickstein2015}
Sohl{-}Dickstein, J., Weiss, E.A., Maheswaranathan, N., Ganguli, S.: Deep
  unsupervised learning using nonequilibrium thermodynamics. In: Bach, F.R.,
  Blei, D.M. (eds.) Proceedings of the 32nd International Conference on Machine
  Learning, {ICML} 2015, Lille, France, 6-11 July 2015. {JMLR} Workshop and
  Conference Proceedings, vol.~37, pp. 2256--2265. JMLR.org (2015)

\bibitem{Sorbo2023}
S{\o}rb{\o}, S., Ruocco, M.: Navigating the metric maze: {A} taxonomy of
  evaluation metrics for anomaly detection in time series. CoRR
  \textbf{abs/2303.01272} (2023)

\bibitem{Su2019}
Su, Y., Zhao, Y., Niu, C., Liu, R., Sun, W., Pei, D.: Robust anomaly detection
  for multivariate time series through stochastic recurrent neural network. In:
  Teredesai, A., Kumar, V., Li, Y., Rosales, R., Terzi, E., Karypis, G. (eds.)
  Proceedings of the 25th {ACM} {SIGKDD} International Conference on Knowledge
  Discovery {\&} Data Mining, {KDD} 2019, Anchorage, AK, USA, August 4-8, 2019.
  pp. 2828--2837. {ACM} (2019)

\bibitem{Tibshirani2001}
Tibshirani, R., Walther, G., Hastie, T.: Estimating the number of clusters in a
  data set via the gap statistic. Journal of the Royal Statistical Society:
  Series B (Statistical Methodology)  \textbf{63}(2),  411--423 (2001)

\bibitem{Tuli2022}
Tuli, S., Casale, G., Jennings, N.R.: Tranad: Deep transformer networks for
  anomaly detection in multivariate time series data. Proc. {VLDB} Endow.
  \textbf{15}(6),  1201--1214 (2022)

\bibitem{Vaswani2017}
Vaswani, A., Shazeer, N., Parmar, N., Uszkoreit, J., Jones, L., Gomez, A.N.,
  Kaiser, L., Polosukhin, I.: Attention is all you need. In: Guyon, I., von
  Luxburg, U., Bengio, S., Wallach, H.M., Fergus, R., Vishwanathan, S.V.N.,
  Garnett, R. (eds.) Advances in Neural Information Processing Systems 30:
  Annual Conference on Neural Information Processing Systems 2017, December
  4-9, 2017, Long Beach, CA, {USA}. pp. 5998--6008 (2017)

\bibitem{Wang2021}
Wang, K., Reimers, N., Gurevych, I.: {TSDAE:} using transformer-based
  sequential denoising auto-encoder for unsupervised sentence embedding
  learning. CoRR  \textbf{abs/2104.06979} (2021)

\bibitem{Wen2022}
Wen, Q., Zhou, T., Zhang, C., Chen, W., Ma, Z., Yan, J., Sun, L.: Transformers
  in time series: {A} survey. CoRR  \textbf{abs/2202.07125} (2022)

\bibitem{Wyatt2022}
Wyatt, J., Leach, A., Schmon, S.M., Willcocks, C.G.: Anoddpm: Anomaly detection
  with denoising diffusion probabilistic models using simplex noise. In:
  {IEEE/CVF} Conference on Computer Vision and Pattern Recognition Workshops,
  {CVPR} Workshops 2022, New Orleans, LA, USA, June 19-20, 2022. pp. 649--655.
  {IEEE} (2022)

\bibitem{Xu2018}
Xu, H., Chen, W., Zhao, N., Li, Z., Bu, J., Li, Z., Liu, Y., Zhao, Y., Pei, D.,
  Feng, Y., Chen, J., Wang, Z., Qiao, H.: Unsupervised anomaly detection via
  variational auto-encoder for seasonal kpis in web applications. In: Champin,
  P., Gandon, F., Lalmas, M., Ipeirotis, P.G. (eds.) Proceedings of the 2018
  World Wide Web Conference on World Wide Web, {WWW} 2018, Lyon, France, April
  23-27, 2018. pp. 187--196. {ACM} (2018)

\bibitem{Xu2022}
Xu, J., Wu, H., Wang, J., Long, M.: Anomaly transformer: Time series anomaly
  detection with association discrepancy. In: The Tenth International
  Conference on Learning Representations, {ICLR} 2022, Virtual Event, April
  25-29, 2022. OpenReview.net (2022)

\bibitem{Yang2022}
Yang, D., Yu, J., Wang, H., Wang, W., Weng, C., Zou, Y., Yu, D.: Diffsound:
  Discrete diffusion model for text-to-sound generation. CoRR
  \textbf{abs/2207.09983} (2022)

\bibitem{Zhang2019}
Zhang, C., Song, D., Chen, Y., Feng, X., Lumezanu, C., Cheng, W., Ni, J., Zong,
  B., Chen, H., Chawla, N.V.: A deep neural network for unsupervised anomaly
  detection and diagnosis in multivariate time series data. In: The
  Thirty-Third {AAAI} Conference on Artificial Intelligence, {AAAI} 2019, The
  Thirty-First Innovative Applications of Artificial Intelligence Conference,
  {IAAI} 2019, The Ninth {AAAI} Symposium on Educational Advances in Artificial
  Intelligence, {EAAI} 2019, Honolulu, Hawaii, USA, January 27 - February 1,
  2019. pp. 1409--1416. {AAAI} Press (2019)

\bibitem{Zhang2023}
Zhang, H., Wang, Z., Wu, Z., Jiang, Y.G.: Diffusionad: Denoising diffusion for
  anomaly detection. arXiv preprint arXiv:2303.08730  (2023)

\bibitem{Zhang2008}
Zhang, R., Zhang, S., Lan, Y., Jiang, J.: Network anomaly detection using one
  class support vector machine. In: Proceedings of the International
  MultiConference of Engineers and Computer Scientists. vol.~1 (2008)

\bibitem{Zhang2007}
Zhang, R., Zhang, S., Muthuraman, S., Jiang, J.: One class support vector
  machine for anomaly detection in the communication network performance data.
  In: Proceedings of the 5th conference on Applied electromagnetics, wireless
  and optical communications. pp. 31--37. Citeseer (2007)

\bibitem{Zhao2020}
Zhao, H., Wang, Y., Duan, J., Huang, C., Cao, D., Tong, Y., Xu, B., Bai, J.,
  Tong, J., Zhang, Q.: Multivariate time-series anomaly detection via graph
  attention network. In: Plant, C., Wang, H., Cuzzocrea, A., Zaniolo, C., Wu,
  X. (eds.) 20th {IEEE} International Conference on Data Mining, {ICDM} 2020,
  Sorrento, Italy, November 17-20, 2020. pp. 841--850. {IEEE} (2020)

\bibitem{Zhou2019}
Zhou, B., Liu, S., Hooi, B., Cheng, X., Ye, J.: Beatgan: Anomalous rhythm
  detection using adversarially generated time series. In: Kraus, S. (ed.)
  Proceedings of the Twenty-Eighth International Joint Conference on Artificial
  Intelligence, {IJCAI} 2019, Macao, China, August 10-16, 2019. pp. 4433--4439.
  ijcai.org (2019)

\bibitem{Zong2018}
Zong, B., Song, Q., Min, M.R., Cheng, W., Lumezanu, C., Cho, D., Chen, H.: Deep
  autoencoding gaussian mixture model for unsupervised anomaly detection. In:
  6th International Conference on Learning Representations, {ICLR} 2018,
  Vancouver, BC, Canada, April 30 - May 3, 2018, Conference Track Proceedings.
  OpenReview.net (2018)

\bibitem{Zonta2020}
Zonta, T., da~Costa, C.A., da~Rosa~Righi, R., de~Lima, M.J., da~Trindade, E.S.,
  Li, G.: Predictive maintenance in the industry 4.0: {A} systematic literature
  review. Comput. Ind. Eng.  \textbf{150},  106889 (2020)

\end{thebibliography}

\vspace{12pt}

\end{document}